\documentclass[10pt,twocolumn,letterpaper]{article}

\usepackage[algorithms]{wacv}

\usepackage{amsmath}
\usepackage{amssymb}
\usepackage{booktabs}
\usepackage{multirow}
\usepackage{pifont}
\usepackage{colortbl}

\newcommand{\green}[1]{{\color{green!60!black}#1}}   %
\renewcommand{\green}[1]{#1}          %

\usepackage{balance}  %

\definecolor{wacvblue}{rgb}{0.21,0.49,0.74}
\usepackage[pagebackref,breaklinks,colorlinks,allcolors=wacvblue]{hyperref}

\def\wacvPaperID{2373}
\def\confName{WACV}
\def\confYear{2027}

\newcommand{\ignore}[1]{}

\makeatletter
\newcommand{\plainfootnote}[1]{\begingroup\renewcommand\thefootnote{}%
  \long\def\@makefntext##1{\noindent##1}\footnotetext{#1}\endgroup}

\makeatother

\title{Explicit Frictional Contact Modeling in Neural Garment Simulation}

\author{%
Thanh-Nguyen Truong\textsuperscript{1}\qquad
Vu-Hoang Tran\textsuperscript{2}\qquad
Ching-Chun Huang\textsuperscript{1,*}\\[0.3em]
\textsuperscript{1}National Yang Ming Chiao Tung University, Taiwan\\
\textsuperscript{2}Ho Chi Minh City University of Technology and Engineering, Vietnam\\[0.3em]
{\small\tt \{thanhnguyenzh.ee13, chingchun\}@nycu.edu.tw, hoangtv@hcmute.edu.vn}
}

\begin{document}
\twocolumn[{
  \renewcommand\twocolumn[1][]{#1}
  \maketitle
  \vspace{-3.2em}
  \begin{center}
  \centering
  \includegraphics[width=0.96\textwidth]{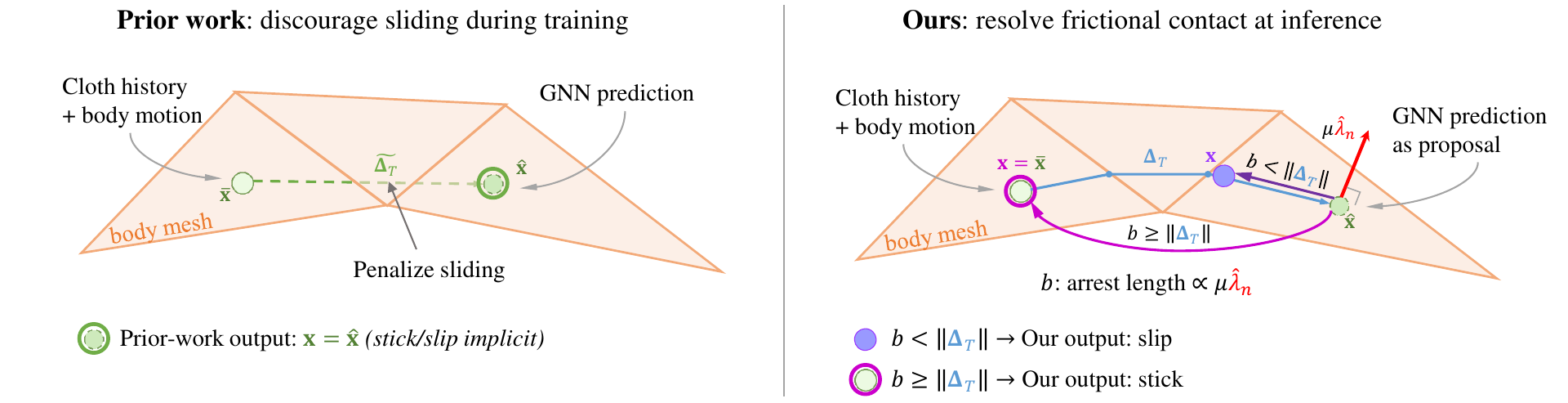}
  \captionsetup{hypcap=false}
  \captionof{figure}{\textbf{Training-time sliding penalty versus explicit inference-time frictional contact.}
  \textbf{Left:} prior work outputs the GNN prediction $\hat{\mathbf{x}}$ and penalizes its approximate tangential displacement $\lVert\widetilde{\boldsymbol{\Delta}}_{T}\rVert$ from the body-carried position $\bar{\mathbf{x}}$ during training, leaving the stick--slip response implicit.
  \textbf{Right:} our operator treats $\hat{\mathbf{x}}$ as a proposal and shortens its tangential slide by a Coulomb arrest length $b$ set by the friction coefficient $\mu$ and the estimated normal load $\widehat{\lambda}_{n}$; the output $\mathbf{x}$ slips if $b$ falls short of the proposed slide and otherwise sticks.}
  \label{fig:problem_surface}
  \end{center}
  \vspace{0.5em}
}]

\begin{abstract}
\plainfootnote{Project page: {\urlstyle{rm}\url{https://trthanhnguyen.github.io/explicit-friction-ngs/}}}%
Graph neural networks (GNNs) simulate garments efficiently through physics-based self-supervision, but leave body--garment friction implicit or inside training energies, with no friction coefficient $\mu$ to set at inference; retraining the tested GNNs with a coefficient-specific friction term barely responds to $\mu$.
Optimization-based solvers decide each contact's stick--slip state from the reaction that governs it, but only by re-optimizing the garment state at every step; a friction term in a one-pass training objective acts as a fixed-weight penalty on sliding, and the trained weights apply no per-contact Coulomb condition at inference.
We therefore resolve friction explicitly after prediction: a Modular Contact Operator, attached to a frozen predictor without retraining or an iterative solve, estimates the normal load, applies a closed-form Coulomb correction along a path across body triangles, and propagates the induced motion through the cloth as a linearized membrane wave.
On HOOD and ContourCraft, it raises sticking and lowers slip monotonically in $\mu$, tracking a reference solver's response while coefficient-specific retraining stays flat, with low penetration and reference-consistent stretching at an eighth of the solver's runtime.
\end{abstract}

\section{Introduction}
\label{sec:intro}

\plainfootnote{$^{*}$Corresponding author.}%
Realistic garment motion is essential to interactive digital humans, virtual try-on, games, and embodied environments, yet high-fidelity cloth simulation remains expensive because deformation, dynamics, and contact must be resolved at every step.
GNN-based neural garment simulators offer an efficient alternative: they predict garment dynamics autoregressively at interactive rates, train from physical objectives without ground-truth trajectories~\cite{santesteban2022snug,bertiche2022ncs,grigorev2023hood}, and generalize across garment topologies and materials.

Recent neural garment simulators improve penetration recovery, multilayer intersections, and self-collision~\cite{grigorev2024contourcraft,liao2024senc}, but these advances address normal geometric feasibility: whether cloth stays unentangled and outside the body.
The tangential response, which decides whether cloth sticks, slips, remains supported, or migrates during motion, stays under-specified.
Classical solvers model it through Coulomb friction, which bounds the tangential reaction by the normal load and the friction coefficient $\mu$; neural simulators instead leave it implicit or penalize relative sliding during training, with $\mu$ fixed as a hyperparameter~\cite{grigorev2023hood}.
Figure~\ref{fig:problem_surface} contrasts that training-time penalty with our inference-time correction.

Two routes to this response exist, and each pays for it differently (Sec.~\ref{sec:background}).
Optimization-based solvers decide each contact's stick--slip state together with the deformation and the normal reaction, by re-optimizing the garment state at every step.
Neural simulators amortize that optimization into a single evaluation of a network, the \emph{predictor}, but a friction term in its training objective is a fixed-weight penalty on sliding: the trained weights apply no per-contact Coulomb condition to the state they produce, and the reaction that decides it is never computed.
We therefore keep the predictor and resolve friction explicitly after it.
A Modular Contact Operator takes the tentative state as a proposal, estimates the available cloth--body normal load, applies a closed-form Coulomb correction along a path across the body mesh, and propagates the induced motion through the surrounding cloth as a linearized membrane wave, while the predictor stays frozen and $\mu$ remains free at inference.

\green{Friction also fits a training objective poorly: it acts only where the garment touches the body, while the network is trained on the garment as a whole (Sec.~\ref{sec:neural_simulation}).
We accordingly place the frictional response outside the training objective.
Our contributions are:}
\begin{itemize}
  \item We introduce an attachable Modular Contact Operator that equips frozen GNN garment predictors with explicit normal-load-dependent Coulomb friction, conditional stick--slip behavior, and inference-time control of $\mu$, without retraining or an iterative cloth solve; the same operator attaches to different predictors unchanged, which we demonstrate on ContourCraft and HOOD.
  \item We show that the operator raises sticking and lowers relative slip monotonically in $\mu$ on both predictors, following a physics-based reference solver, while keeping body penetration low and the stretching trend consistent with the reference at $8.1\times$ lower end-to-end cost.
  \item We validate the operator through cumulative component ablations that isolate Geometry-Aware Contact Traversal and Friction-Induced Motion Propagation, and probe the normal-load estimate against closed-form loads.
\end{itemize}

\section{Related work}
\label{sec:related_work}

\subsection{Contact and friction in cloth solvers}
\label{sec:solver_contact}

Classical cloth simulators couple deformation with collision processing and impulse-based or constrained contact response~\cite{baraff1998large,bridson2002robust}.
Velocity-space, projective-dynamics, and IPC-family methods enforce normal-load-dependent Coulomb friction and conditional stick--slip behavior through iterative contact solves~\cite{daviet2015fast,li2018argus,daviet2020simple,ly2020projective,li2020ipc,li2021cipc}.
Recent work improves robustness and speed or differentiates these solvers for inverse problems~\cite{ando2024cubic,zeng2025fast,zheng2026barrierfree,li2022diffcloth,zeng2026gradients}.
These formulations provide the behavior we seek, but revise deformation and contact variables iteratively at each step, so their per-frame cost stays far from interactive rates (Sec.~\ref{sec:runtime}).

\subsection{Learned garment animation and draping}
\label{sec:learned_animation}

Early learned garment models predict pose-dependent corrections or wrinkles through skinning, pose-space deformation, virtual bones, or neural fields~\cite{santesteban2019learning,patel2020tailornet,pan2022virtualbones,zhang2022motion}.
Physical self-supervision removes the need for solver trajectories in garment draping~\cite{bertiche2021pbns,deluigi2023drapenet}.
These methods efficiently reproduce appearance or equilibrium, but body--garment interaction is learned from data or represented by normal collision constraints rather than an explicit dynamic friction law, so whether a garment sticks or slips is neither controllable nor bounded by the normal load.

\subsection{Self-supervised neural garment dynamics}
\label{sec:self_supervised_dynamics}

MeshGraphNets established autoregressive graph-based simulation from solver-supervised trajectories~\cite{pfaff2021mgn}.
SNUG and NCS instead train garment dynamics with physical objectives, but neither includes tangential body contact~\cite{santesteban2022snug,bertiche2022ncs}.
HOOD generalized this self-supervised setting across garments and topologies~\cite{grigorev2023hood}, and introduced a sliding-work objective proportional to $\mu m g\cos\phi$ and tangential displacement~\cite{grigorev2023hood,brown2018accurate,geilinger2020add}: a gravity-based load proxy that penalizes sliding without a conditional sticking response and fixes $\mu$ during training.
ContourCraft and SENC improve multilayer intersections and self-collision, while EUNet learns the fabric's constitutive dissipation; none of them changes this body--garment friction model~\cite{grigorev2024contourcraft,liao2024senc,shao2024eunet}.

Consequently, prior neural garment simulators do not provide an explicit normal-load-dependent Coulomb bound with conditional stick--slip behavior and freely variable inference-time $\mu$; their tangential response is fixed at training time, and Sec.~\ref{sec:quantitative_results} shows that it barely moves even under coefficient-specific retraining.

\section{Background and problem formulation}
\label{sec:background}

\subsection{Optimization-based frictional contact}
\label{sec:coulomb_contact}

For cloth--body contact $i$, let $\mathbf{u}_{i}=(u_{N,i},\mathbf{u}_{T,i})$ be the relative contact velocity, $\lambda_{n,i}$ the compressive normal reaction, and $\mathbf f_{T,i}$ the tangential force, all defined over the same discrete time step.
The Signorini conditions include $\lambda_{n,i}\geq 0$, while Coulomb friction imposes $\lVert\mathbf f_{T,i}\rVert\leq\mu\lambda_{n,i}$~\cite{daviet2020simple}; the bound is contact-local because $\lambda_{n,i}$ varies across vertices.
A sticking contact has $\mathbf{u}_{T,i}=\mathbf{0}$ with tangential force inside the cone; a sliding contact lies on its boundary and opposes motion.

Optimization-based simulators obtain these states by minimizing the incremental potential of one implicit step over the state $\mathbf x$ of the current step.
We denote its inertia, elastic, gravity, and contact energies, in method-dependent forms, by $E_{\mathrm{phys}}(\mathbf x;\mathbf s^{t})$ given the recent states and body motion $\mathbf s^{t}$; a friction potential completes the objective.
\mbox{In IPC's formulation~\cite{li2020ipc},}
\begin{align}
\mathbf{x}^{t+1}&=\operatorname*{argmin}_{\mathbf{x}}\,E_{\mathrm{phys}}(\mathbf{x};\mathbf{s}^{t})+D_{\mu}(\mathbf{x}),\notag\\
D_{\mu}(\mathbf{x})&=\sum_{i\in\mathcal C}\mu\,\lambda_{n,i}\,f_{0}\!\left(\lVert\boldsymbol{\Delta}_{T,i}(\mathbf{x})\rVert\right),
\label{eq:solver_program}
\end{align}
where $\mathcal C$ is the active contact set, $\boldsymbol{\Delta}_{T,i}(\mathbf x)=h\,\mathbf u_{T,i}$ is the relative tangential displacement over the step of length $h$, and $f_0$ is a mollifier of the sliding distance that rounds its kink at zero.
The friction term is weighted by the reaction itself: $\lambda_{n,i}$ and the tangent frame are lagged from the previous solve, refreshed from the new state, and the minimization is repeated until the step's residual is met, so each contact's stick--slip state is decided by the reaction that governs it.
The price is an iterative solve per step, with energy, derivative, and linear-system evaluations at every iterate, repeated across the lag updates~\cite{li2018argus,ly2020projective,li2020ipc}.
The later cubic-barrier solver of Ando~\cite{ando2024cubic}, which we call PPF after its released implementation and adopt as our reference, keeps this pattern of per-step state optimization; Sec.~\ref{sec:runtime} measures its cost against ours.

\subsection{Self-supervised neural garment simulation}
\label{sec:neural_simulation}

Self-supervised neural garment simulators instead learn an autoregressive transition without solver-generated target trajectories.
A graph network $G_{\theta}$ communicates deformation along cloth edges and body interaction through proximity relations~\cite{pfaff2021mgn,grigorev2023hood,grigorev2024contourcraft,liao2024senc}, and training minimizes the incremental potential at its prediction, in expectation over training states~\cite{santesteban2022snug,grigorev2023hood}:
\begin{align}
\hat{\mathbf{x}}^{t+1}&=G_{\theta}(\mathbf{s}^{t}),\notag\\
\theta^{\star}&=\operatorname*{argmin}_{\theta}\,\mathbb{E}\big[E_{\mathrm{phys}}(\hat{\mathbf{x}}^{t+1};\mathbf{s}^{t})+E_{\mathrm{friction}}\big],\notag\\
E_{\mathrm{friction}}&=\sum_{i\in\mathcal C}\mu\,m_{i}g\cos\phi_{i}\,\lVert\widetilde{\boldsymbol{\Delta}}_{T,i}(\hat{\mathbf{x}}^{t+1})\rVert,
\label{eq:gnn_program}
\end{align}
where the friction term, introduced by HOOD and retained by later predictors of this family, penalizes the approximate tangential displacement $\widetilde{\boldsymbol{\Delta}}_{T,i}$ from the body-carried position with a gravity-based load proxy $m_{i}g\cos\phi_{i}$, $m_{i}$ being the vertex mass and $\phi_{i}$ the local body slope~\cite{grigorev2023hood}.
At inference, this amortization replaces the per-step minimization with a single network evaluation, enabling fast autoregressive rollout.

Friction, however, enters this program only as a penalty whose weight $\mu m_{i}g\cos\phi_{i}$ is a training constant: the term has no parameters of its own, $\mu$ is fixed, and the load is a proxy the network never computes.
Descent on $\theta$ therefore sees friction as a fixed-weight cost on sliding, traded against the other energies in expectation over the data, and the resulting $\theta^{\star}$ is one vector shared by every contact and every training state.
\green{The trade falls unevenly on friction.}
\green{The elastic and inertial energies are costs that every cloth element contributes to, whereas the Coulomb condition applies only to the contacting fraction of the garment.}
\green{Because the network predicts the garment as a whole, it can still lower the expected objective while leaving that fraction's friction unresolved.}
\green{Furthermore, because the fabric is coupled, the loading that friction induces in the surrounding cloth is likewise absent from the prediction (Sec.~\ref{sec:quantitative_results}).}
At inference $G_{\theta^{\star}}$ is evaluated once: nothing compares the state it produces with the Coulomb bound, and no stick--slip state is decided per contact.

\subsection{Explicit contact modeling after prediction}
\label{sec:predict_then_correct}

Per-step optimization buys the stick--slip decision with an iterative solve; one-pass prediction removes the solve and, with it, the decision.
\green{Rather than enlarging the training objective, our} design keeps the fast pass and restores the decision after it,
\begin{equation}
\hat{\mathbf{x}}^{t+1}=G_{\theta^{\star}}(\mathbf{s}^{t}),
\qquad
\mathbf{x}^{t+1}=P_{\mu}(\hat{\mathbf{x}}^{t+1}),
\label{eq:predict_correct}
\end{equation}
where the frozen predictor supplies a tentative state and our Modular Contact Operator $P_{\mu}$ corrects it with a runtime coefficient $\mu$.
The operator must estimate the reaction the predictor never computes, apply the bound to each contact, keep the corrected endpoint along the body surface, and carry the corrected motion into the surrounding fabric.
Section~\ref{sec:method} constructs these four stages; each runs once per step, so the stick--slip state is decided without iteratively minimizing Eq.~\eqref{eq:solver_program} to stationarity and without retraining the predictor (Sec.~\ref{sec:runtime}).

\section{Modular contact operator}
\label{sec:method}

\subsection{Overview}
\label{sec:method_overview}

\begin{figure*}[t]
  \centering
  \includegraphics[width=\textwidth]{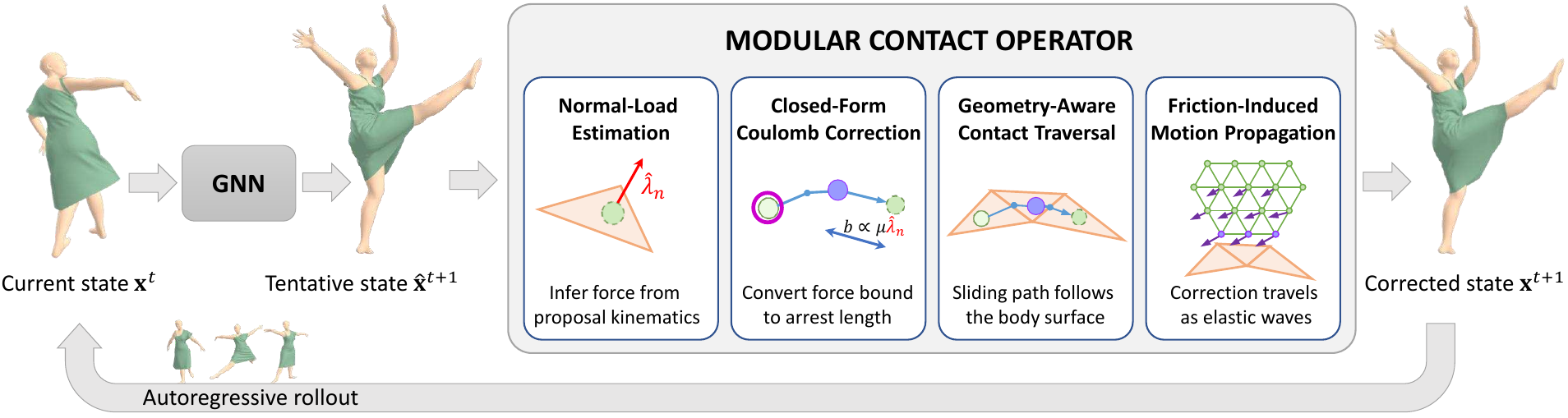}
  \caption{\textbf{Pipeline of the Modular Contact Operator.}
  A frozen GNN maps the current autoregressive state to a tentative state; the four stages correct it in the order shown, and the corrected state re-enters the rollout history for the next prediction.}
  \label{fig:operator_pipeline}
\end{figure*}

The neural predictor first emits a tentative state $\hat{\mathbf{x}}^{t+1}$, after which our operator produces the corrected state $\mathbf{x}^{t+1}=P_{\mu}(\hat{\mathbf{x}}^{t+1})$ (Fig.~\ref{fig:operator_pipeline}).
We form the proximity contact set $\mathcal C_\varepsilon=\{i\mid d_i<\varepsilon\}$, where $d_i$ is the signed body distance of garment vertex $i$ at the tentative state and $\varepsilon$ a fixed shell width; for each $i\in\mathcal C_\varepsilon$, the closest body face supplies a local normal--tangent frame and the body motion needed for relative displacement.
Within this frame, each stage answers one requirement of a per-contact Coulomb correction without a solve: the operator (i) estimates an effective compressive normal load from the proposal kinematics, (ii) shortens the predicted sliding by a Coulomb-bounded arrest length, leaving stick or residual slip, (iii) places the corrected endpoint along the body surface by Geometry-Aware Contact Traversal, and (iv) carries the correction through the surrounding fabric by Friction-Induced Motion Propagation.
Outside $\mathcal C_\varepsilon$ the prediction is retained except where the propagation stage moves the surrounding fabric (Sec.~\ref{sec:propagation}); at $\mu=0$ every vertex retains the baseline prediction exactly.

\subsection{Normal-Load Estimation}
\label{sec:normal_load}

The Coulomb correction requires a normal-load estimate $\widehat{\lambda}_{n}$, but the neural predictor emits accelerations rather than contact reactions.
We therefore infer a step-averaged load from the proposal's departure from gravitational free fall, following impulse-based cloth contact, where the tangential correction is bounded by the normal velocity change that collision response produces~\cite{bridson2002robust}; here the neural proposal supplies that change without an inner contact solve.
We drop the vertex index $i$ until Sec.~\ref{sec:propagation}.
For time step $h$, let $\mathbf x^{t-1}$ and $\mathbf x^{t}$ be the vertex's two preceding positions and $\hat{\mathbf x}^{t+1}$ its tentative proposal; let $m$ be its lumped mass, $g$ the gravity magnitude, $\mathbf e_{\mathrm{up}}$ the upward unit vector, and $\mathbf n$ the outward body normal.
The estimator reads the acceleration the proposal implies, forms a gravity-compensated force residual, and retains its compressive normal component:
\begin{equation}
\begin{aligned}
\widehat{\mathbf a}
&=\frac{\hat{\mathbf x}^{t+1}-2\mathbf x^{t}+\mathbf x^{t-1}}{h^{2}},\\[2pt]
\widehat{\mathbf f}^{\,\mathrm{res}}
&=m\left(\widehat{\mathbf a}+g\mathbf e_{\mathrm{up}}\right),\\[2pt]
\widehat{\lambda}_{n}
&=\left[\mathbf n^{\mathsf T}\widehat{\mathbf f}^{\,\mathrm{res}}\right]_{+}
=m\left[g\,\mathbf n^{\mathsf T}\mathbf e_{\mathrm{up}}
+\mathbf n^{\mathsf T}\widehat{\mathbf a}\right]_{+},
\end{aligned}
\label{eq:normal_load}
\end{equation}
where $[q]_{+}=\max(q,0)$ retains compression and excludes adhesion; the estimate has units of step-averaged force and requires no barrier-stiffness calibration.
We emphasize that $\widehat{\lambda}_{n}$ is not the exact contact reaction of the coupled cloth--body system; it is a kinematic estimate that sets the available friction budget.

The scalar form exposes two complementary behaviors.
The projected gravity term $g\,\mathbf n^{\mathsf T}\mathbf e_{\mathrm{up}}$ provides a \emph{gravitational response} that assigns the static-support load according to surface orientation.
The normal acceleration term $\mathbf n^{\mathsf T}\widehat{\mathbf a}$ provides an \emph{impact response}: for a locally fixed normal, a normal velocity change $\Delta v_N$ over one step gives $\mathbf n^{\mathsf T}\widehat{\mathbf a}=\Delta v_N/h$, so $m\,\mathbf n^{\mathsf T}\widehat{\mathbf a}$ is the corresponding step-averaged normal force.
Thus ballistic free fall cancels the gravity contribution, whereas arresting incoming normal motion raises the estimated load in proportion to its deceleration.
Section~\ref{sec:coulomb_correction} converts this force to an arrest length; the supplement probes both responses against closed-form loads.

\subsection{Closed-Form Coulomb Correction}
\label{sec:coulomb_correction}
\setcounter{topnumber}{1}

Over one step, let $\bar{\mathbf x}^{t+1}$ be the position to which the body's motion alone would carry the vertex, and $\boldsymbol{\Delta}_{T}$ the tangent-plane projection of the remaining relative displacement $\hat{\mathbf x}^{t+1}-\bar{\mathbf x}^{t+1}$.
Because the operator acts on positions rather than forces, we convert the Coulomb force bound of Sec.~\ref{sec:coulomb_contact} into the maximum relative tangential displacement it can cancel within one step.
Assuming the maximum opposing friction impulse $\mu\widehat{\lambda}_{n}h$ acts over the step, this displacement is
\begin{equation}
b
=\frac{\mu\widehat{\lambda}_{n}h^{2}}{m}
=\mu h^{2}\left[\mathbf n^{\mathsf T}\!\left(\widehat{\mathbf a}+g\mathbf e_{\mathrm{up}}\right)\right]_{+},
\label{eq:arrest_length}
\end{equation}
which we call the arrest length.
The lumped mass cancels, so $b$ is a length determined only by $\mu$, the step, and the proposal's compressive normal departure from free fall.

With static-band parameter $\epsilon_v$, the correction is the block soft-threshold
\begin{equation}
\boldsymbol{\Delta}^{+}_{T}
=\max\!\left(
0,
1-
\frac{b}{\sqrt{\lVert\boldsymbol{\Delta}_{T}\rVert^2+(\epsilon_vh)^2}}
\right)
\boldsymbol{\Delta}_{T},
\label{eq:coulomb_correction}
\end{equation}
the classical velocity-level friction update~\cite{bridson2002robust} written in displacement units, with $\epsilon_vh$ regularizing the static--sliding transition as barrier-based frictional contact does with a velocity scale of its own~\cite{li2020ipc}. We set $\epsilon_v=2\times10^{-3}\,\mathrm{m\,s^{-1}}$.
Thus the operator produces a sticking response when the arrest length exceeds the regularized tangential displacement in the denominator; otherwise the remaining displacement is interpreted as slip, up to the band's smoothing near the transition.
This local closed form decides stick or slip without a contact solve, and allows runtime $\mu$ control without asking the GNN to predict the contact reaction.

\subsection{Geometry-Aware Contact Traversal}
\label{sec:contact_traversal}

\begin{figure}[t]
  \centering
  \includegraphics[width=\columnwidth]{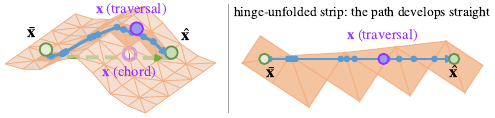}
  \caption{\textbf{Geometry-Aware Contact Traversal.}
  \textbf{Left:} from the same endpoints $\bar{\mathbf{x}}$ and $\hat{\mathbf{x}}$ on a folded mesh, the ambient chord (green) passes through the surface and lands $\mathbf{x}$ (chord) inside the body, whereas our triangle walk (blue) follows the crossed faces and keeps $\mathbf{x}$ (traversal) at its contact clearance.
  \textbf{Right:} unfolding those faces about their hinges develops the walk into a straight segment.}
  \label{fig:contact_traversal}
\end{figure}

Applying the tangential shrink along a straight ambient-space chord can cut through a curved body or use a stale normal across a crease.
We instead reconstruct the slide path between its two known free-space endpoints, $\bar{\mathbf x}^{t+1}$ and $\hat{\mathbf x}^{t+1}$: unlike cross-triangle continuation for mesh embeddings~\cite{shen2020phongsurface,shao2024splattingavatar}, the path exists only for the current correction and must preserve metric length for the Coulomb update.
After projecting both endpoints to their supporting face planes while retaining their signed plane gaps, the walk starts at the body-carried foot and repeats a project--clip--unfold step: it aims at the tentative endpoint developed into the current face, advances to the first face boundary, unfolds the remaining target across the crossed edge by a hinge rotation, and lifts the junction off the crease along the bisector of the two face normals.
Because each unfold is isometric, the resulting polyline is straight through the developed face strip and preserves surface arc length (Fig.~\ref{fig:contact_traversal}); it is not claimed to be a globally shortest geodesic, and the crossed faces themselves set the path discretization.

Each crossed face $k$ carries its own arrest length $b_k$, obtained by re-projecting the residual $\widehat{\mathbf f}^{\,\mathrm{res}}$ of Sec.~\ref{sec:normal_load} onto that face normal.
In the path's arc-length coordinate $\sigma\in[0,L]$, the endpoint stops where the remaining slide equals the local arrest length,
\begin{equation}
L-\sigma^{\star}=b_{k(\sigma^{\star})},
\label{eq:traversal_landing}
\end{equation}
with $\sigma^{\star}$ the landing position and $k(\sigma)$ the containing segment; we regularize this relation by the static band of Sec.~\ref{sec:coulomb_correction} and select the nearest admissible landing from the tentative end, so $b_k=0$ retains the tentative endpoint while an arrest length spanning the path returns the vertex to $\bar{\mathbf x}^{t+1}$.
The load and normal are refreshed on the landing face, a normal-completion step restores the pre-correction local gap subject to a non-penetration clamp, and ambiguous or degenerate walks revert to the chord landing.
Section~\ref{sec:ablations} measures the effect of this stage.

\subsection{Friction-Induced Motion Propagation}
\label{sec:propagation}

\begin{figure}[t]
  \centering
  \includegraphics[width=\columnwidth]{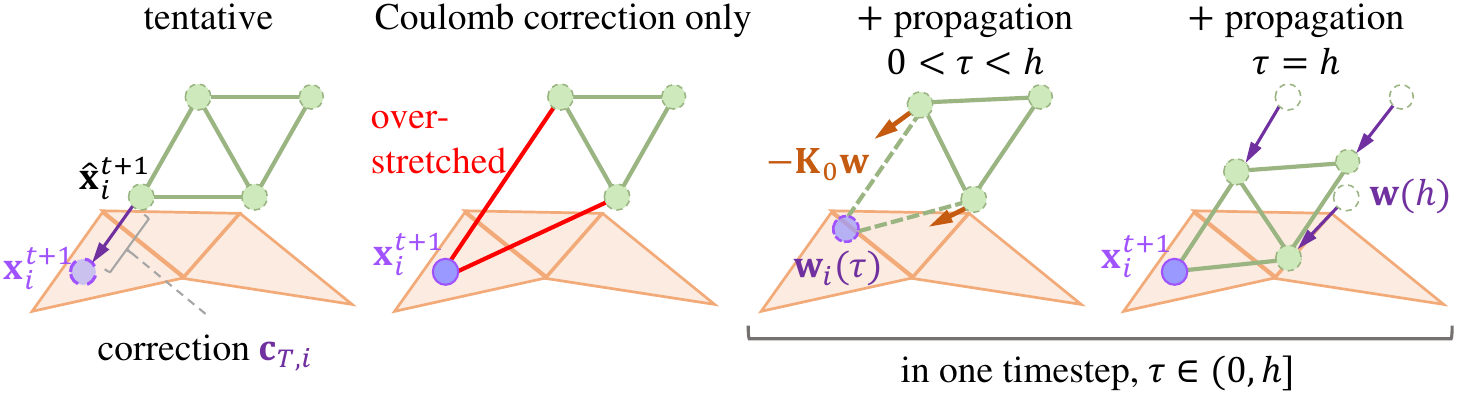}
  \caption{\textbf{Friction-Induced Motion Propagation} within one frame.
  Left to right: the tentative state with the tangential correction $\mathbf c_{T,i}$ of contact $i$; the Coulomb correction alone, which over-stretches the edges to the uncorrected neighbours; propagation at $0<\tau<h$, where the contact follows the ramp $\mathbf w_i(\tau)$ and the elastic pull $-\mathbf K_0\mathbf w$ carries the surrounding fabric; and $\tau=h$, where the neighbours have moved by $\mathbf w(h)$ and no displacement discontinuity remains.}
  \label{fig:propagation}
\end{figure}

The corrections of Secs.~\ref{sec:normal_load}--\ref{sec:contact_traversal} move only the contact vertices while the surrounding fabric still follows the tentative prediction, and the resulting discontinuity over-stretches the fabric at the edge of the contact set.
Our operator's final stage, Friction-Induced Motion Propagation, removes that discontinuity at its source by carrying each contact's tangential correction $\mathbf c_{T,i}$---the displacement from tentative position to corrected tangential landing---through the surrounding fabric with a linearized membrane dynamics model; the transported motion is a material-dependent wave (Fig.~\ref{fig:propagation}).

Over one frame we integrate a conservative, mass-lumped membrane wave for an incremental displacement field $\mathbf w$,
\begin{equation}
\mathbf M\,\ddot{\mathbf w}=-\mathbf K_0\,\mathbf w,
\qquad \tau\in(0,h],
\label{eq:propagation_wave}
\end{equation}
where $\tau$ is time within the frame, $\mathbf M$ the lumped mass matrix, and $\mathbf K_0$ the garment's membrane stiffness linearized at the tentative state, so the garment's own material sets how far the correction travels within the frame.
Each corrected contact is driven to its correction by the fixed ramp $\mathbf w_i(\tau)=s(\tau/h)\,\mathbf c_{T,i}$ with $s(r)=3r^2-2r^3$, which starts and ends at rest; contacts that received no correction, and pinned vertices, are held at zero, and the free vertices receive $\hat{\mathbf x}^{t+1}_i+\mathbf w_i(h)$.
To keep the transported motion from stretching the fabric it crosses, we limit the added field's per-edge strain to a budget $\gamma_{\max}$, $\lVert\mathbf w_i-\mathbf w_j\rVert\le\gamma_{\max}\,\ell_{ij}$ with $\ell_{ij}$ the edge length at the tentative state~\cite{provot1995deformation,goldenthal2007efficient,ando2024cubic}, splitting each edge's clip between its endpoints by inverse mass~\cite{bridson2002robust}; Sec.~\ref{sec:ablations} shows the consequence of removing the limit.
Because the transported motion can also carry free vertices toward the body, a final clamp restores any body gap the wave reduced.
At $\mu=0$ the correction set is empty and the full operator, propagation included, reduces to the identity.

\section{Experimental results}
\label{sec:evaluation}

We test whether the Modular Contact Operator produces controlled stick--slip behavior consistent with a reference solver, transfers across HOOD and ContourCraft checkpoints, and preserves garment quality; we then isolate traversal, propagation, and strain limiting and report runtime.

\begin{figure*}[t]
  \centering
  \includegraphics[width=\textwidth]{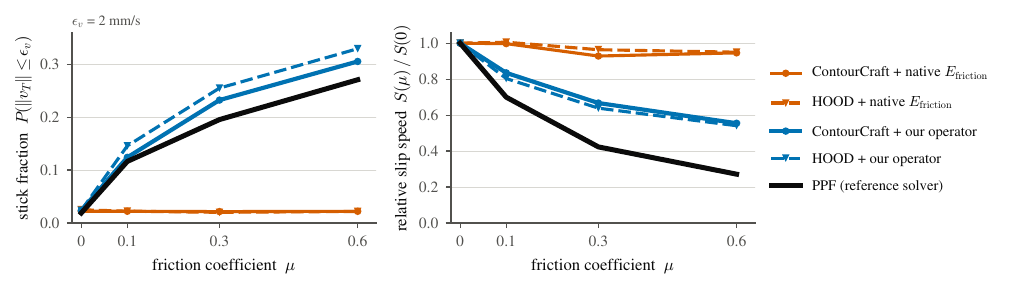}
  \caption{\textbf{$\mu$-dependent frictional response} over the $40$ matched sequences.
  \textbf{Left:} pooled mass--time stick fraction at the operator's static band $\epsilon_v$ (Sec.~\ref{sec:coulomb_correction}).
  \textbf{Right:} pooled mass--time relative slip speed, each candidate normalized by its own $\mu=0$ rollout.
  $E_{\mathrm{friction}}$ curves are coefficient-specific retrainings.}
  \label{fig:friction_response}
\end{figure*}

\begin{figure*}[t]
  \centering
  \includegraphics[width=0.8\textwidth]{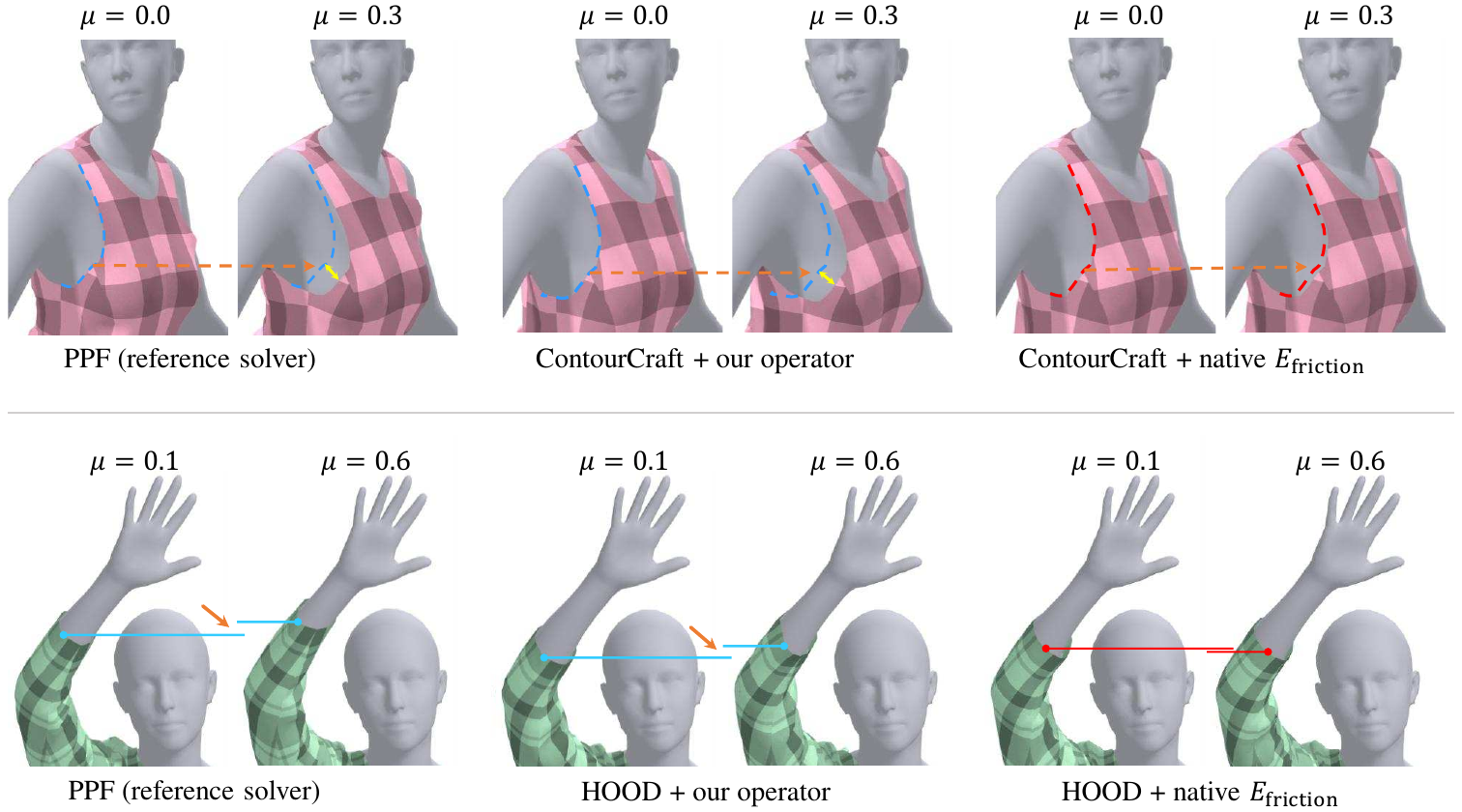}
  \caption{\textbf{Qualitative $\mu$-dependent frictional response.}
  Top: tank-top armhole after a dance motion at $\mu\in\{0,0.3\}$ on ContourCraft; bottom: long-sleeve cuff during an arm raise at $\mu\in\{0.1,0.6\}$ on HOOD.
  Each group shows PPF, the backbone with our operator, and coefficient-specific $E_{\mathrm{friction}}$ retraining, lower $\mu$ at left.}
  \label{fig:qualitative}
\end{figure*}

\subsection{Experimental setup}
\label{sec:experimental_setup}
\setcounter{topnumber}{2}

We evaluate two GNN garment predictors, HOOD~\cite{grigorev2023hood} and ContourCraft~\cite{grigorev2024contourcraft}, in two configurations.
\emph{$+$ our operator} attaches our operator at inference to a backbone we retrain with no friction term, so its frictional response cannot be inherited from training.
\emph{$+$ native $E_{\mathrm{friction}}$} retrains the same architecture with its inherited friction energy and runs without our operator.
This energy is the friction term $E_{\mathrm{friction}}$ of Eq.~\eqref{eq:gnn_program}~\cite{grigorev2023hood}.
Its coefficient is fixed during training, so we retrain one checkpoint per $\mu$.
PPF~\cite{ando2024cubic} is the physics-based reference solver.

We use $\mu\in[0,0.6]$, following ARGUS~\cite{li2018argus} and spanning friction coefficients reported for real textiles~\cite{inoue2011surface}, on four garment templates from HOOD and ten AMASS CMU motions~\cite{mahmood2019amass}, for $40$ matched garment--motion sequences.
Five motions come from the SNUG/HOOD evaluation sets and five from a pre-registered extension spanning sustained-load cases and a predicted low-load null; the grid retains sequences completed by PPF at every coefficient.
All neural rollouts use $h=1/30$ s, and within each comparison garment, motion, coefficient, evaluation interval, and metric implementation are fixed.
The supplement lists the sequences, material mapping, operator settings, and solver parameters.

\subsection{Dynamic garment evaluation}
\label{sec:dynamic_evaluation}

We quantify the $\mu$ response on the matched grid, then visualize it in the same rollouts and on unseen garments.

\subsubsection{Quantitative results}
\label{sec:quantitative_results}

\begin{figure*}[t]
  \centering
  \includegraphics[width=0.8\textwidth]{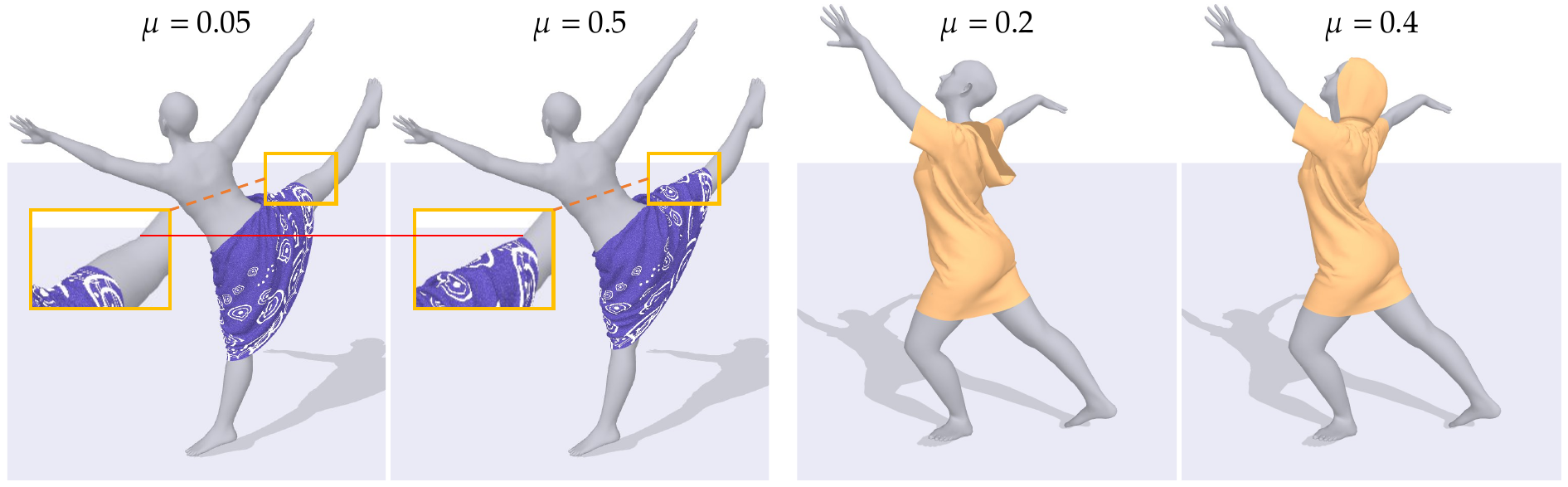}
  \caption{\textbf{Inference-time control of the friction coefficient.}
  Left: a long skirt over an arabesque at $\mu\in\{0.05,0.50\}$, with insets enlarging the skirt--leg contact.
  Right: a hooded dress at $\mu\in\{0.20,0.40\}$.
  All rollouts use our operator on one retrained ContourCraft backbone; the garments and motions were absent from its training.}
  \label{fig:showoff}
\end{figure*}

Figure~\ref{fig:friction_response} reports two contact metrics pooled over contact mass and time across all $40$ sequences.
The \emph{stick fraction} is the fraction of contacting cloth at or below the operator's static band $\epsilon_v=2$ mm/s (Sec.~\ref{sec:coulomb_correction}).
The \emph{relative slip speed} $S(\mu)/S(0)$ is the mean tangential cloth--body speed normalized by the same candidate's $\mu=0$ rollout.

Despite coefficient-specific retraining, both native $E_{\mathrm{friction}}$ candidates remain nearly flat.
In contrast, our operator increases sticking and reduces slip monotonically on both frozen backbones: across the coefficient range, its stick fraction rises alongside PPF's, and its slip falls in the reference direction, although less steeply.
The shared trend across HOOD and ContourCraft supports transfer of the explicit response between the two tested architectures.

\begin{table}[t]
  \centering
  \caption{\textbf{Garment-quality checks over the matched grid.}
  Mean penetration: fraction of garment vertex-frames with negative signed body distance, averaged over $\mu\in\{0.1,0.3,0.6\}$ and the $40$ sequences; the lower mean within each backbone is bolded.
  Stretching-energy ratio: each candidate's frame-mean $E_{\mathrm{stretching}}$ at $\mu$ relative to its $\mu=0$ rollout, averaged over the $40$ sequences; {\setlength{\fboxsep}{1.5pt}\colorbox{gray!25}{shaded}} rows increase monotonically with $\mu$, as PPF's does.}
  \label{tab:quality}
  \resizebox{\columnwidth}{!}{%
  \begin{tabular}{lcccc}
    \toprule
    \multirow{2}{*}{Method} & Mean pen.\,$\downarrow$ & \multicolumn{3}{c}{$E_{\mathrm{str}}(\mu)/E_{\mathrm{str}}(0)$} \\
    \cmidrule(lr){3-5}
    & [$10^{-3}\,\%$] & $\mu=0.1$ & $\mu=0.3$ & $\mu=0.6$ \\
    \midrule
    PPF (reference solver) & 0.2 & \cellcolor{gray!25}1.004 & \cellcolor{gray!25}1.038 & \cellcolor{gray!25}1.314 \\
    \midrule
    ContourCraft $+$ our operator & \textbf{6.1} & \cellcolor{gray!25}0.994 & \cellcolor{gray!25}1.047 & \cellcolor{gray!25}1.139 \\
    ContourCraft $+$ native $E_{\mathrm{friction}}$ & 10.4 & 0.927 & 0.927 & 0.924 \\
    \midrule
    HOOD $+$ our operator & \textbf{8.6} & \cellcolor{gray!25}1.003 & \cellcolor{gray!25}1.040 & \cellcolor{gray!25}1.073 \\
    HOOD $+$ native $E_{\mathrm{friction}}$ & 13.9 & 1.004 & 0.994 & 1.029 \\
    \bottomrule
  \end{tabular}}
\end{table}

Table~\ref{tab:quality} shows lower penetration for the operator than for native $E_{\mathrm{friction}}$ retraining on each backbone.
Because sticking contact transmits tangential reaction into the fabric~\cite{ly2020projective}, stretching should rise with $\mu$: the operator's ratio does, as PPF's does, whereas the native $E_{\mathrm{friction}}$ candidates remain approximately flat.

\subsubsection{Qualitative results}
\label{sec:qualitative_results}

Figure~\ref{fig:qualitative} compares PPF, our operator, and native friction-energy training on the same rollouts.
In the top row, raising $\mu$ from $0$ to $0.3$ retains more motion-induced displacement and leaves the tank-top armhole farther forward on the torso; in the bottom row, raising $\mu$ from $0.1$ to $0.6$ lets the rising forearm carry the cuff upward instead of letting it slip down.
In both, our operator follows the direction and approximate extent of PPF, whereas friction-energy training changes weakly at top and moves the cuff the opposite way at bottom.

Figure~\ref{fig:showoff} isolates inference-time control on one retrained ContourCraft backbone, with garments and motions absent from its training: raising $\mu$ from $0.05$ to $0.50$ keeps more of the skirt supported by the raised leg, and the hood that slips off at $\mu=0.20$ stays on the head at $\mu=0.40$.
These examples show control at user-specified coefficients rather than reference agreement.

\subsection{Ablation study}
\label{sec:ablations}

\begin{figure}[t]
  \centering
  \includegraphics[width=\columnwidth]{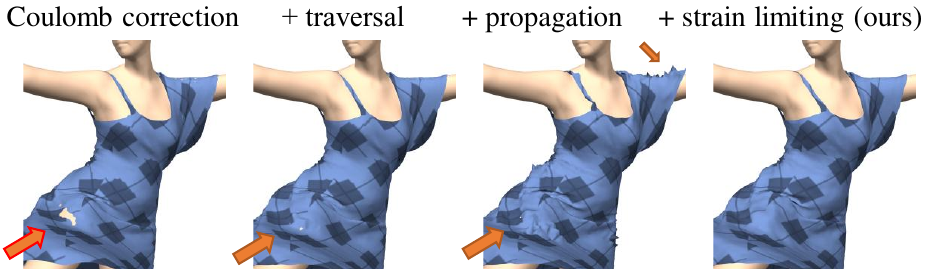}
  \caption{\textbf{Qualitative component ablation}, ContourCraft backbone, $\mu=0.3$; columns follow the rows of Table~\ref{tab:components}.
  Under the same body motion, Coulomb correction alone drives cloth through the thigh (arrow); traversal closes the crossing but leaves the fabric taut; unlimited propagation gives unstable, excessive stretching; strain limiting removes both.}
  \label{fig:ablation_qualitative}
\end{figure}

Table~\ref{tab:components} begins with normal-load estimation and Coulomb correction, then adds traversal, propagation, and strain limiting in execution order, holding the neural backbone, proximity contact set, checkpoint, initialization, evaluation settings, and $\mu=0$ anchor fixed across the rows; Fig.~\ref{fig:ablation_qualitative} shows the same ladder on one frame.
Every row uses the load estimate of Sec.~\ref{sec:normal_load}, which the supplement probes separately against closed-form loads on prescribed scenes.

\begin{table}[t]
  \centering
  \caption{\textbf{Cumulative component ablations.}
  Starting from normal-load estimation and Coulomb correction, subsequent rows add traversal, propagation, and strain limiting in execution order on the retrained ContourCraft backbone with checkpoint, contact set, and $\mu=0$ anchor shared.
  Mean penetration: fraction of garment vertex-frames with negative signed body distance over the $40$ sequences and $\mu\in\{0.1,0.3,0.6\}$.
  Deviation to reference: per-sequence $100\,|r/r_{\mathrm{ref}}-1|$ between the row's and PPF's stretching-energy ratios, mean (median).
  Bold: lowest mean per column.}
  \label{tab:components}
  \setlength{\tabcolsep}{4pt}
  \resizebox{\columnwidth}{!}{%
  \begin{tabular}{@{}lcccc@{}}
    \toprule
    \multirow{2}{*}{Configuration} & Mean pen.\,$\downarrow$ & \multicolumn{3}{c}{Deviation to reference [\%]\,$\downarrow$} \\
    \cmidrule(lr){3-5}
    & [$10^{-3}\,\%$] & $\mu=0.1$ & $\mu=0.3$ & $\mu=0.6$ \\
    \midrule
    Coulomb correction & 11.5 & 2.9 (1.6) & 13.3 (10.5) & 38.6 (27.5) \\
    \quad $+$ traversal & 6.5 & \textbf{2.3} (1.4) & 12.6 (9.5) & 36.8 (26.5) \\
    \quad $+$ propagation & 9.2 & 3.0 (1.4) & 1149.0 (16.5) & 1268.6 (114.1) \\
    \quad $+$ strain limiting (ours) & \textbf{6.1} & 2.8 (1.1) & \textbf{6.9} (4.3) & \textbf{17.3} (11.1) \\
    \bottomrule
  \end{tabular}}
\end{table}

\paragraph{Geometry-Aware Contact Traversal.}
The first row runs the normal-load estimate and the Coulomb correction alone, placing each corrected endpoint along the straight ambient chord; the second adds the traversal, which walks the endpoint across the body surface and completes it in the normal direction under a non-penetration clamp (Sec.~\ref{sec:contact_traversal}).
The stage nearly halves the mean body-penetration rate while leaving the stretching-energy deviation from the reference essentially unchanged: the geometric placement reduces penetration at no measurable cost in garment loading.

\paragraph{Friction-Induced Motion Propagation.}
The last two rows add, in turn, tangential-correction propagation and its per-edge strain limit (Sec.~\ref{sec:propagation}).
Because both the target discontinuity and the risk introduced concern fabric strain, we compare their stretching-energy deviation from the reference.
At $\mu=0.1$, where the frictional correction remains small, all four rows remain closely comparable.
At $\mu=0.3$ and $0.6$, propagation without strain limiting raises the mean and median deviations well above those of the no-propagation row and visibly over-stretches fabric beside the contact set.
With strain limiting, the complete operator instead yields the lowest mean and median deviations at both coefficients.
Transporting the correction therefore preserves reference-consistent loading only when the added field is strain-limited.

\subsection{Runtime}
\label{sec:runtime}

At $\mu=0.6$, the operator's busiest evaluated coefficient, our method produces a sequence in $27.2$ s versus $220.4$ s for PPF on average: $8.1\times$ faster end-to-end, including PPF's fixed settling prologue, and $5.6\times$ on marginal per-frame cost, $86$ ms against $481$ ms per output frame.
Averaged per output frame, the GNN forward pass takes $58\,\%$ and the operator $42\,\%$, of which propagation is $13\,\%$; the supplement reports the timing scope and breakdown.
The explicit operator thus retains much of the predictor's speed advantage, while the coupled reference solve recovers a stronger frictional response.

\section{Conclusion}
\label{sec:conclusion}

Optimization-based solvers decide each contact's stick--slip state from its reaction but re-optimize the garment state at every step; one-pass predictors remove the solve, and with friction reduced to a fixed-weight training penalty, the tested backbones remain nearly insensitive to the coefficient.
We therefore keep the one-pass predictor and make the Coulomb law explicit after it.
The Modular Contact Operator gives frozen neural garment predictors normal-load-dependent Coulomb friction with runtime control of $\mu$, resolving each contact's stick--slip response in closed form on the tentative state and carrying the induced motion through the surrounding cloth.
Across the ContourCraft and HOOD backbones it produces a $\mu$-dependent frictional response while keeping penetration low and the stretching trend consistent with the reference solver.

\paragraph{Limitations.}
The operator remains a one-pass body--garment approximation rather than a globally converged solve.
It does not model cloth--cloth friction, guarantee collision-free trajectories, or recover membrane-tension-induced normal traction such as hoop loading in tight garments.
The wave transports corrected motion without resolving reactions; recovering coupled elastic contact forces would require a local solve or force supervision.
\green{Our results argue for placing the frictional response outside the training objective in the tested setting, not against learning to model friction: a friction objective or training procedure in which the reaction, and with it the stick--slip decision, depends on the network's parameters---through a predicted reaction or a short unrolled solve---could move part of this response into the weights, and we regard that as a future direction.}

\section*{Acknowledgments}
This work was financially supported in part (project number: 112UA10019) by the Co-creation Platform of the Industry Academia Innovation School, NYCU, under the framework of the National Key Fields Industry-University Cooperation and Skilled Personnel Training Act, from the Ministry of Education (MOE) and industry partners in Taiwan.
It is also supported in part by the National Science and Technology Council, Taiwan, under Grant NSTC-115-2634-F-A49-011, Grant NSTC-114-2218-E-A49-024, Grant NSTC-115-2425-H-A49-001, Grant NSTC-114-2622-E-A49-027, Grant NSTC-115-2221-E-A49-124-MY3, Grant NSTC-115-2218-E-A49-017 and in part by the Higher Education Sprout Project of the National Yang Ming Chiao Tung University and the Ministry of Education (MOE), Taiwan.
It is also partly supported by MediaTek Inc., Hon Hai Research Institute, E.SUN Bank, and Industrial Technology Research Institute.
(Corresponding author: Ching-Chun Huang.)
During this research, \mbox{Thanh-Nguyen} Truong was kindly supported by the \mbox{TaiwanICDF} \mbox{Higher Education Scholarship Program}.

\balance
{
  \small
  \bibliographystyle{ieeenat_fullname}
  \bibliography{main}
}

\end{document}